%% file: main.tex
\documentclass[10pt]{article}
\usepackage[preprint]{tmlr}

\input{math_commands.tex}

\usepackage{amsmath}
\usepackage{amssymb}
\usepackage{booktabs}
\usepackage{graphicx}
\usepackage{array}
\usepackage{xcolor}
\usepackage{microtype}
\usepackage{hyperref}
\usepackage{url}

\title{NitroGen: Observations on the Action Dynamics, Visual Instruction Following, and Spatial Exploration}

\author{\name Shengyuan Lin \email shengyu3@andrew.cmu.edu \\
      \addr Carnegie Mellon University
      \AND
      \name Zhiling Li \email zhilingl@andrew.cmu.edu \\
      \addr Carnegie Mellon University
      \AND
      \name Haoyu Nie \email hnie@andrew.cmu.edu \\
      \addr Carnegie Mellon University
      \AND
      \name Ozan K. Tonguz \email tonguz@andrew.cmu.edu \\
      \addr Carnegie Mellon University}

\begin{document}

\maketitle

\begin{abstract}
NitroGen is an open vision-action foundation model for game control. It is trained by behavior cloning on online gameplay videos. In this paper, we study the frozen controller in \textit{Hollow Knight} and \textit{Trackmania} as examples along three dimensions: action adaptation, visual instruction following, and spatial exploration. In five independent six-minute trials, the action distribution changes across game contexts. Horizontal movement drops from 53.4\% in the initial corridor to 41.4\% at a later scene that requires jumping. Jump actions rise from 11.5\% to 24.5\%, which suggests that NitroGen responds to changes in the local visual state, but it still fails to finish the jump scene. Visual prompts provide only weak steering. Graphical arrows affect behavior more than text prompts, but neither prompt type produces reliable instruction following. Visual-state clustering shows concentrated visitation. In the full additional dataset, NitroGen visits more clusters than Random but fewer than Human on average. Its mean entropy is lower than Random's. The Human--NitroGen entropy ordering depends on the cluster count. Our results indicate that NitroGen can adapt its actions to local context, but it lacks the planning, recovery, and exploration needed for longer-horizon game control.
\end{abstract}

\section{Introduction}

Generalist game-playing agents are a useful setting for studying pixel-based control. The agent must read the screen and output actions directly. It also needs to recover from mistakes and continue toward a goal over time. Prior work has studied game agents with reinforcement learning in Atari and StarCraft II~\citep{mnih2015dqn,badia2020agent57,vinyals2019alphastar}, LLM/API-based control in systems such as Voyager~\citep{wang2023voyager}, and embodied-agent benchmarks such as MineDojo~\citep{fan2022minedojo}.

NitroGen is an open vision-to-action foundation model for game control, jointly developed by researchers from NVIDIA, Stanford University, and the California Institute of Technology (Caltech). It maps raw RGB frames to gamepad actions without access to the game-specific APIs or internal game states. The released 500M-parameter controller uses a DiT vision-action architecture and is trained by behavior cloning on roughly 40,000 hours of public human gameplay videos from more than 1,000 games~\citep{magne2026nitrogen}. It has a universal simulator wrapper that provides a common interface for commercial game binaries. Action patterns are learned during pretraining. We freeze those learned weights during every evaluation. NitroGen is designed as a fast reactive controller, not as a complete long-horizon planning system.

We study the capabilities and limitations of this frozen controller. The main question is whether NitroGen only produces local reflexes, or whether its behavior can support longer game progress. This matters for inference-time interventions. A high-level module may detect that the agent is stuck and propose a macro-action. A visual prompt may add an arrow or text cue to the screen. However, these methods can help only if the frozen controller can respond to the new signal and execute the intended action.

We therefore focus on the controller without adding any new planning module. A short-term action predictor may produce reasonable actions in a local scene, but still fail as a game-playing agent. Game progress also requires goal maintenance, recovery, navigation, and exploration. If the controller cannot reliably perform the needed low-level behavior, then better high-level advice may not lead to better play.

We evaluate NitroGen with three analyses. We call them D1, D2, and D3, respectively. D1 tests whether its action distribution changes across game contexts. D2 tests whether visual prompts can steer its behavior. D3 measures how much of the game space it explores compared with random and human baselines. Our results show that the three studies can distinguish local reactive control from longer-horizon failure.

\section{Related Work}

\paragraph{Game-playing agents and generalization.}

Game environments have long been used as benchmarks for assessing agents. The Arcade Learning Environment created a shared platform for evaluating general agents~\citep{bellemare2013ale}. DQN and Agent57 showed strong reinforcement-learning performance in Atari, while AlphaStar did so in StarCraft II~\citep{mnih2015dqn,badia2020agent57,vinyals2019alphastar}. Procgen shifted the focus from fixed-game performance to generalization on generated levels~\citep{cobbe2020procgen}. Voyager depends on language-model planning together with executable skills. Gato trains one sequence model across multiple embodied domains~\citep{wang2023voyager,reed2022gato}. These systems center on task performance, planning, or cross-domain behavior. They do not directly test how a frozen vision-action controller responds when it must recover from local failure without reward updates.

\paragraph{Learning game control from large-scale demonstrations.}

MineRL and MineDojo offer large-scale datasets for Minecraft agents, including human demonstrations and online knowledge~\citep{guss2019minerl,fan2022minedojo}. Video PreTraining (VPT) applies behavior cloning on online video to learn game-control priors~\citep{baker2022vpt}. NitroGen follows a similar path. It is a generalist pixel-to-gamepad model trained on gameplay videos from more than 1,000 games~\citep{magne2026nitrogen}. However, behavior cloning can suffer from covariate shift and compounding errors during closed-loop rollout~\citep{ross2011dagger}, and imitation policies can exhibit causal confusion~\citep{dehaan2019causal}. We study the released frozen controller to determine whether its local action changes are enough for instruction following, obstacle completion, and exploration.

\paragraph{Visuomotor conditioning.}

Some visuomotor policies use task or language conditioning during training. BC-Z, RT-1, RT-2, VIMA, and language-conditioned imitation learning include task descriptions, language, or multimodal commands as part of the policy input~\citep{jang2022bcz,brohan2022rt1,brohan2023rt2,jiang2023vima,lynch2020language}. Our NitroGen setting is different. The evaluated controller receives only pixels. Text and arrows placed on the screen are therefore visual changes to the observation, not native command inputs.

\paragraph{Planning and exploration.}

Some agent systems divide low-level control from higher-level planning. SayCan combines language planning with learned affordances, while MuZero and DreamerV3 use learned models for search or latent prediction~\citep{ahn2022saycan,schrittwieser2020muzero,hafner2023dreamerv3}. Other research looks at exploration. DIAYN, Go-Explore, and Plan2Explore encourage agents to discover diverse skills or states~\citep{eysenbach2018diayn,ecoffet2021goexplore,sekar2020plan2explore}. Our work does not introduce a new planner or exploration objective. It provides a diagnostic evaluation of NitroGen. The model shifts its local action distribution when the visual context changes, yet this does not guarantee obstacle completion, prompt controllability, or broad state coverage.

\section{Problem Formulation}

We study NitroGen as a frozen reactive controller that maps game frames to gamepad actions. In our implementation, the action representation consists of a 21-dimensional button vector plus left and right joystick values. The model output is converted into gamepad input through thresholding and joystick scaling. This interface is suitable for fast reactive motor control, but it does not expose an explicit memory, goal representation, symbolic state, or text-instruction channel.

We focus on three measurable properties. First, we measure whether NitroGen changes its action distribution as Hollow Knight progresses from a corridor that requires horizontal navigation to an obstacle that requires jumping. Second, if visual prompting is to be useful, the model must treat an overlaid arrow or instruction as a control-relevant cue rather than as irrelevant image content. Third, an agent that is supposed to play games rather than only produce plausible controller traces should visit a sufficiently broad set of visual states.

\begin{table}[htbp]
\centering
\caption{NitroGen overview and evaluation-relevant caveat.}
\label{tab:nitrogen_overview}
\vspace{1mm}
\begin{tabular}{@{}p{0.28\linewidth}p{0.62\linewidth}@{}}
\toprule
Attribute & Description \\
\midrule
Training source & 40,000 hours of gameplay videos \\
Game coverage & 1,000 games \\
Model scale & 500M-parameter DiT \\
Control scope & Pixel observations to gamepad actions; the released 500M model should be treated as a fast-reacting System-1 controller rather than a long-horizon end-to-end player. \\
\bottomrule
\end{tabular}
\end{table}

The central evaluation question is whether NitroGen's failures can be characterized as stable action- and state-level patterns rather than isolated task failures. We organize the problem into three measurable questions:

\begin{enumerate}
    \item \textbf{Context-dependent action adaptation:} Does NitroGen change its control strategy when gameplay moves from horizontal navigation to a jump-dependent obstacle, and is the change sufficient to make progress?
    \item \textbf{Instruction blindness:} Can NitroGen use visual or textual cues embedded in the input frame to change its movement direction?
    \item \textbf{Exploration constraints:} Does NitroGen cover a broad visual state space, or does it remain confined to a small region?
\end{enumerate}

\begin{table}[htbp]
\centering
\caption{Organization of Experimental Results and Evidence Chain.}
\label{tab:evidence_chain}
\vspace{1mm}
\resizebox{0.95\linewidth}{!}{%
\begin{tabular}{@{}lll@{}}
\toprule
Study & Failure mode & Conclusion \\
\midrule
D1 & Local adaptation & Horizontal control decreases and jumping increases, but the obstacle remains uncleared \\
D2 & Semantic blindness & Limited, modality and environment-dependent prompt effect \\
D3 & Spatial confinement & Concentrated visitation; coverage ordering depends on the setting \\
\bottomrule
\end{tabular}
}
\end{table}

\begin{table}[htbp]
\centering
\caption{Experimental scope. The study is diagnostic; all rows use the frozen NitroGen controller without weight updates.}
\label{tab:experimental_scope}
\footnotesize
\vspace{1mm}
\resizebox{0.95\linewidth}{!}{%
\begin{tabular}{@{}p{0.06\linewidth}p{0.26\linewidth}p{0.24\linewidth}p{0.28\linewidth}@{}}
\toprule
Study & Time horizon & Setting & Measured outcome \\
\midrule
D1 & Five six-minute Hollow Knight trials; windows at $0$--$1$ s, $0$--$60$ s, and $300$--$360$ s & Corridor followed by a jump puzzle & Horizontal-control activity, jump frequency, direction switching, and sustained direction states \\
D2 & $n=100$ short trials per condition & No prompt, right-arrow prompt, and ``Go Right'' text prompt & Mean right/left movement and $P(R>L)$ \\
D3 & Fifteen-minute open-space trajectories sampled at 1 FPS & NitroGen, Random, and Human play & Visited clusters, entropy, PCA projection, bootstrap entropy distribution \\
\bottomrule
\end{tabular}
}
\end{table}

This setup is used to test the capabilities of NitroGen, not to improve it during inference. The tested prompts are limited to arrows and short text cues. They should not be treated as the full set of possible prompts. Human gameplay is used as a proxy ground truth because NitroGen is trained from human gameplay videos, but it is not treated as a unique optimal policy. Random play is used only as a diagnostic baseline.

We note that Hollow Knight and Trackmania are stress tests, not representative benchmarks for all games. Hollow Knight exposes horizontal navigation, multi-step jumps, and recovery after failed jumps. Trackmania exposes fast directional control. We do not establish how often either game, or a similar control regime, appeared in NitroGen pretraining. Distribution shift may contribute to the measured behavior. Our claims concern the tested games and the released frozen controller.

We use three measurements to test the related limits. D1 compares the action distribution across two Hollow Knight scenes. D2 adds arrows or text to the input frames and measures whether the controller changes its behavior. D3 clusters the observed frames and compares the state coverage of NitroGen, random play, and human play. The studies test action choice, response to visual cues, and exploration.

The three studies distinguish local reaction from longer game progress. We observe that NitroGen can change its action distribution when the scene changes. However, it does not mean that it can clear the jump, follow prompts reliably, or cover a large part of the environment. This observation matters before adding a planner or a VLM module. Even if such a module selects a better next step, the final gamepad action still has to come from the frozen controller.

\section{Approach}

\subsection{Context-Dependent Action Evaluation}

We ran five Hollow Knight trials from the same initial corridor. Each trial used the frozen NitroGen checkpoint. Each trial contains 21,600 environment frames at 60 frames per second (FPS), equal to six minutes of game time. Model inference pauses the game between environment steps. Wall-clock inference latency is therefore excluded. Before each trial, we reset the model and manually returned the game to the same initial state.

We use three windows: Immediate ($0$--$1$ s), Early ($0$--$60$ s), and Late ($300$--$360$ s). Early is the main corridor window. Late is the later obstacle window. The corridor is deliberately constrained. The main local requirement is horizontal movement to the left or right. The later scene is a jump puzzle that blocks the next scene. Passing it requires roughly four continuous jumps with horizontal control. The Early and Late comparison is a context comparison, not a pure time comparison.

At frame $t$, let $x_t$ denote the normalized horizontal value of the left joystick. We call a frame horizontally active when $|x_t| > 0.25$. We use the same threshold in every trial and window. Let
\[
M_t =
\begin{cases}
1, & \text{if } |x_t| > 0.25,\\
0, & \text{otherwise}.
\end{cases}
\]
The horizontal-active value in Table~\ref{tab:d1_context_shift} is the average of $M_t$ over all frames in the corresponding window. Jump activity is the fraction of frames where the \texttt{SOUTH} button is pressed. We also report direction switch rate and the longest continuous left, neutral, or right state. Persistence is not a failure by itself. We use these metrics only to describe the temporal shape of the behavior. All aggregate values are means across the five trials. The paired trial trajectories are shown in the figure.

\subsection{Visual Instruction-Following Evaluation}

We evaluate three prompt conditions: no prompt, a right-arrow image overlay, and the text instruction ``Go right. Move the character to the right.'' Each condition has 100 trials. We measure the fraction of right and left movement. Then we report mean right movement, mean left movement, and $P(\text{right}>\text{left})$.

The prompt study asks whether a rightward cue increases rightward movement. We test both an arrow prompt and a text prompt. An arrow is a visual cue. Text requires language understanding from pixels. The simple prompts set a low bar for a control signal. The test does not rank prompt designs. It asks whether a clear on-screen cue can act as a dependable basic command. NitroGen is not trained with a language-conditioned action interface, so weak text response is expected. Language-conditioned imitation learning includes language during policy training~\citep{lynch2020language}.

For trial $i$, let $r_i$ and $\ell_i$ be the fraction of measured frames with rightward and leftward movement. $P(R>L)$ is the fraction of trials where $r_i>\ell_i$. Mean right and left proportions measure action frequency. $P(R>L)$ measures trial-level directional dominance.

\subsection{Spatial Exploration Evaluation}

To measure visual state-space coverage, we use fifteen-minute open-space trajectories from NitroGen, Random, and Human play. D3 differs from D1. The agent is not restricted to a corridor and a jump obstacle. Broad coverage requires both horizontal navigation and jumping over a longer horizon. We sample frames at 1 FPS, embed them, and compute state-visitation entropy:
The embedding pipeline is
\[
\begin{gathered}
256{\times}256\text{ frames at 1 FPS}
\rightarrow \text{SigLIP }(\text{1024}d)\\
\rightarrow \text{PCA }(64d)
\rightarrow k\text{-means }(K=50)\\
\rightarrow \text{Shannon entropy }H.
\end{gathered}
\]
We use SigLIP as a frozen perceptual embedding model for offline visual-state analysis~\citep{zhai2023siglip}. Each frame is assigned to a cluster, and we compute
\[
H = -\sum_{i=1}^{K} p_i \log p_i.
\]
Higher entropy means broader and more balanced visual-state coverage. Clustered visual entropy is only a diagnostic approximation. It is not a full geometry-aware exploration objective. Entropy-based exploration in continuous state spaces depends on the geometry of the representation~\citep{guo2021gem}. We compare NitroGen with Random and Human baselines. We use frame-level bootstrap sampling with $n=1000$ to estimate uncertainty within the original recordings.

The metric is not a direct score of game progress. It measures visual confinement. Random play may visit states that do not help winning. Human play may visit states through goal-directed progress. Since NitroGen is trained from human gameplay videos, we treat human gameplay as a close ground truth for purposeful state coverage, not as an exact optimal policy. If NitroGen covers fewer states than Random under the same embedding and clustering pipeline, then the frozen controller is spatially conservative at the level of observed visual states.

\subsection{Interpretation Criteria}

We use relative thresholds. In D1, the main evidence is paired change across repeated trials. In the corridor scene, we expect more horizontal movement while more jumping should be tried at the later obstacle. A consistent action shift supports visual-context sensitivity, but it is not sufficient evidence of successful play. In D1, successful long-horizon execution requires clearing the jump obstacle.

\begin{table}[htbp]
\centering
\caption{Interpretation of the main diagnostic metrics.}
\label{tab:metric_interpretation}
\vspace{1mm}
\footnotesize
\begin{tabular}{@{}p{0.20\linewidth}p{0.34\linewidth}p{0.34\linewidth}@{}}
\toprule
Metric & High value means & Low value means \\
\midrule
Horizontal activity & More frames contain substantial left/right joystick input & Horizontal control is used less often \\
Jump frequency & Jump is selected more often in the window & Jump is selected less often in the window \\
Direction-switch rate & Direction state changes frequently & Longer sustained direction states \\
$P(R>L)$ & Rightward movement dominates more trials & Rightward movement dominates fewer trials \\
State entropy & Broader or more balanced visual-state coverage & Concentrated visitation in fewer visual clusters \\
\bottomrule
\end{tabular}
\end{table}

For D1, all five paired trials show less horizontal activity and more jumping in the Late window. None clears the obstacle. For D2, even the best graphical prompt gives an incomplete shift. The arrow suppresses leftward movement in Hollow Knight but increases rightward movement in Trackmania. Prompt response is environment-dependent. It is not a reliable semantic channel. The original D3 recordings place NitroGen below both baselines. The additional runs change this ordering (Appendix~\ref{app:d3}).

Note that the claims of our paper are limited to the tested frozen controller, prompt forms, and games. We do not claim that all visual prompting fails. Similarly, we do not claim that VLM-based systems cannot help. Future NitroGen checkpoints may behave differently. The practical question is which failure mode should be addressed next.

Clearly, no single metric is sufficient for drawing a conclusion. Although a higher jump rate at a jump obstacle is locally appropriate, it does not necessarily demonstrate planning. Low state entropy can also be acceptable if an agent follows a narrow successful path. The important pattern is the combination of adaptation to local action, failure to complete the obstacle, weak instruction response, and low state coverage relative to baselines.

\section{Numerical Results}

\subsection{Context-Dependent Action Adaptation}

\begin{figure}[htbp]
    \centering
    \begin{minipage}[t]{0.48\linewidth}
        \centering
        \IfFileExists{fig/d1_corridor.png}{\includegraphics[width=\linewidth]{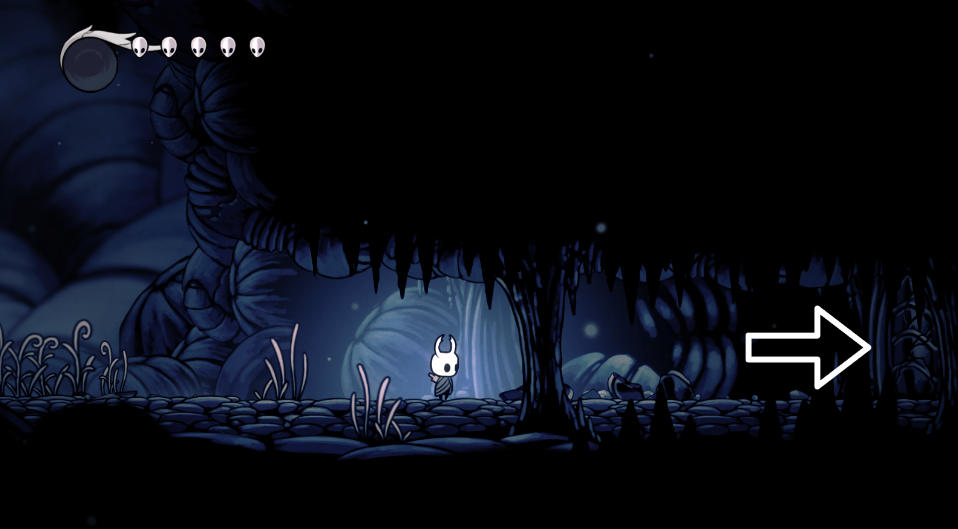}}{\fbox{\parbox[c][0.24\textheight][c]{0.92\linewidth}{\centering Add initial corridor screenshot\\as \texttt{fig/d1\_corridor.png}}}}
        \\(a) Initial navigation corridor
    \end{minipage}\hfill
    \begin{minipage}[t]{0.48\linewidth}
        \centering
        \IfFileExists{fig/d1_jump_obstacle.png}{\includegraphics[width=\linewidth]{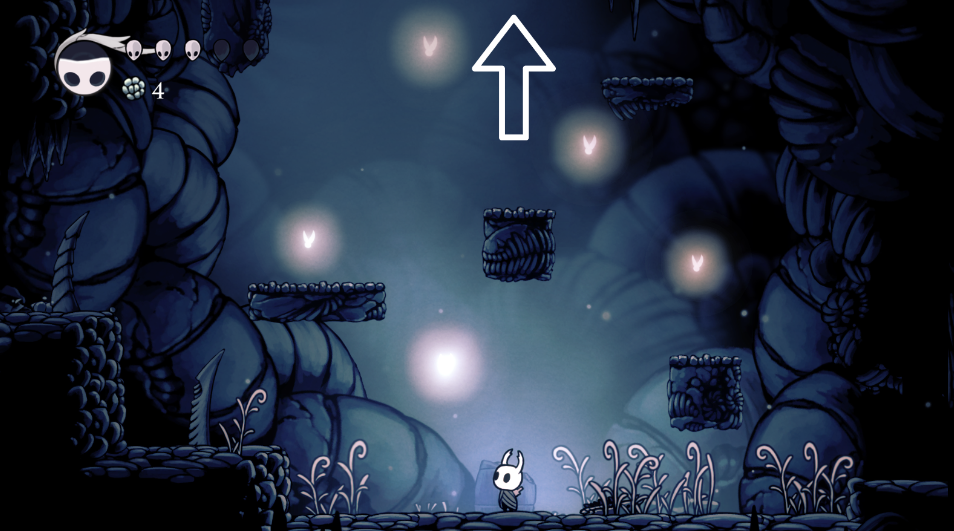}}{\fbox{\parbox[c][0.24\textheight][c]{0.92\linewidth}{\centering Add jump-obstacle screenshot\\as \texttt{fig/d1\_jump\_obstacle.png}}}}
        \\(b) Later jump-dependent obstacle
    \end{minipage}
    \caption{Representative Hollow Knight contexts for D1. The Early window contains corridor navigation; the Late window contains a jump puzzle that requires roughly four coordinated jumps with horizontal control before the agent can enter the next scene.}
    \label{fig:d1_contexts}
\end{figure}

\begin{table}[htbp]
\centering
\caption{D1 metrics across five independent Hollow Knight trials. Values are mean $\pm$ standard deviation across trials.}
\label{tab:d1_context_shift}
\vspace{1mm}
\footnotesize
\resizebox{0.95\linewidth}{!}{%
\begin{tabular}{@{}lcccc@{}}
\toprule
Window & Horizontal active & Jump pressed & Direction switches & Longest direction run (s) \\
\midrule
Immediate ($0$--$1$ s) & $67.3\pm15.7\%$ & $24.0\pm21.5\%$ & $5.76\pm4.42\%$ & $0.50\pm0.18$ \\
Early ($0$--$60$ s) & $53.4\pm3.9\%$ & $11.5\pm2.4\%$ & $3.67\pm0.40\%$ & $3.45\pm0.78$ \\
Late ($300$--$360$ s) & $41.4\pm8.4\%$ & $24.5\pm2.8\%$ & $3.20\pm0.33\%$ & $5.50\pm2.12$ \\
\bottomrule
\end{tabular}
}
\end{table}

\begin{figure}[htbp]
    \centering
    \includegraphics[width=0.92\linewidth]{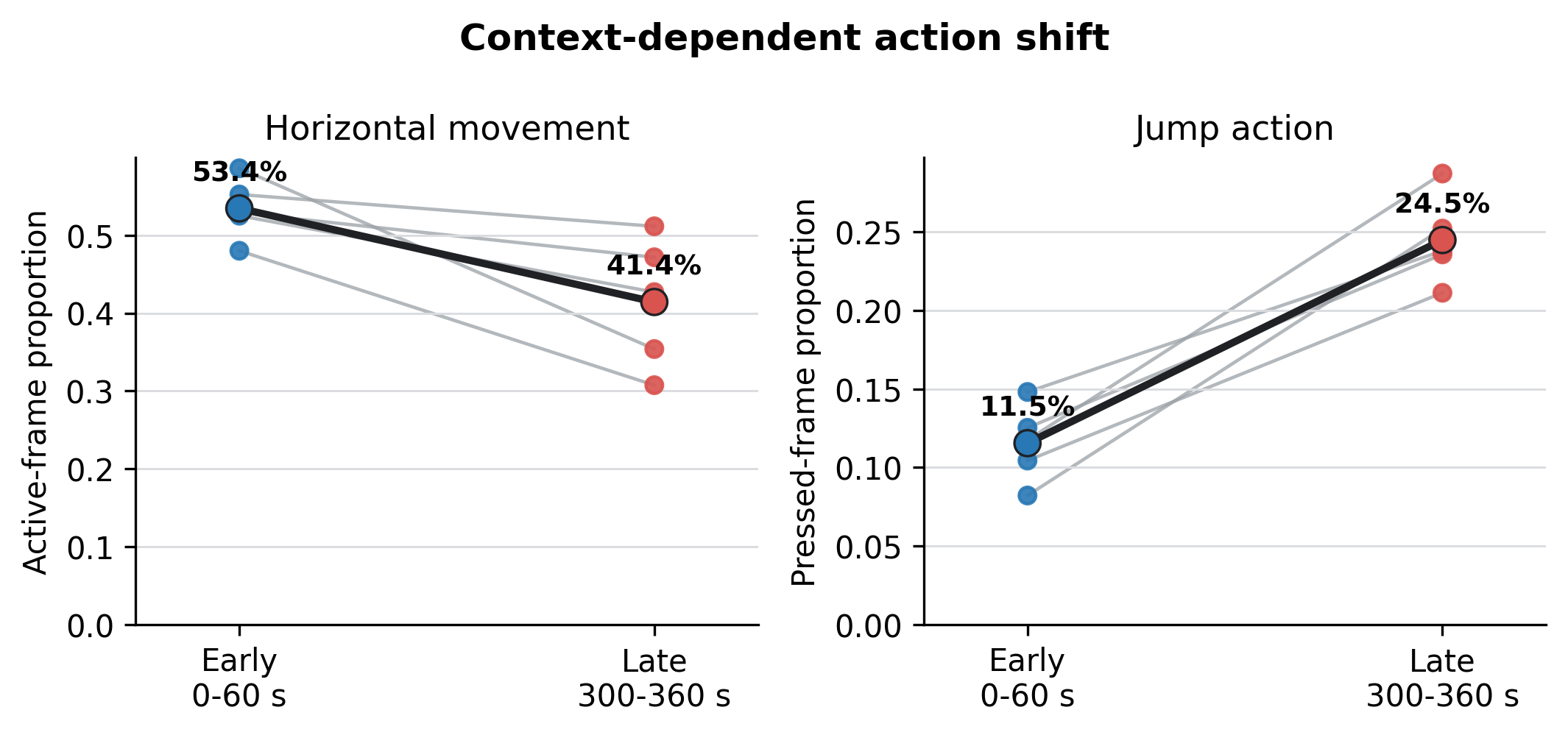}
    \caption{Paired action-distribution changes from the Early corridor to the Late jump-dependent obstacle. Thin lines show individual trials and thick lines show means. Horizontal activity decreases in all five trials, while jump frequency increases in all five trials.}
    \label{fig:d1_strategy_shift}
\end{figure}

Table~\ref{tab:d1_context_shift} reports the D1 action metrics. Fig.~\ref{fig:d1_contexts} shows the two contexts. Fig.~\ref{fig:d1_strategy_shift} shows the paired changes across trials. Horizontal activity drops from $53.4\%$ in Early to $41.4\%$ in Late. Jump frequency rises from $11.5\%$ to $24.5\%$. All five paired trials move in the same direction. The controller uses more horizontal movement in the corridor. It uses more jumping at the obstacle.

The higher jump rate is locally appropriate at the obstacle, but it is not sufficient for obstacle completion. NitroGen never clears the obstacle. In several runs, it starts the jump sequence, then reverses direction before finishing the path. It returns toward earlier platforms instead of preserving the multi-step objective. We do not claim direct evidence of an internal memory module. The observed failure is a long-horizon execution and recovery failure. It is consistent with weak memory or planning over the full jump sequence. Direction switching also falls from $3.67\%$ to $3.20\%$. The longest sustained direction state grows from $3.45$ to $5.50$ seconds. The later behavior is more sustained, but not uniformly locked. NitroGen can choose a relevant local action type. It does not organize those actions into a successful obstacle-completion strategy. Additional D1 results are reported in Appendix~\ref{app:d1}.

\subsection{Visual Instruction Following}

\begin{figure}[htbp]
    \centering
    \includegraphics[width=0.5\linewidth]{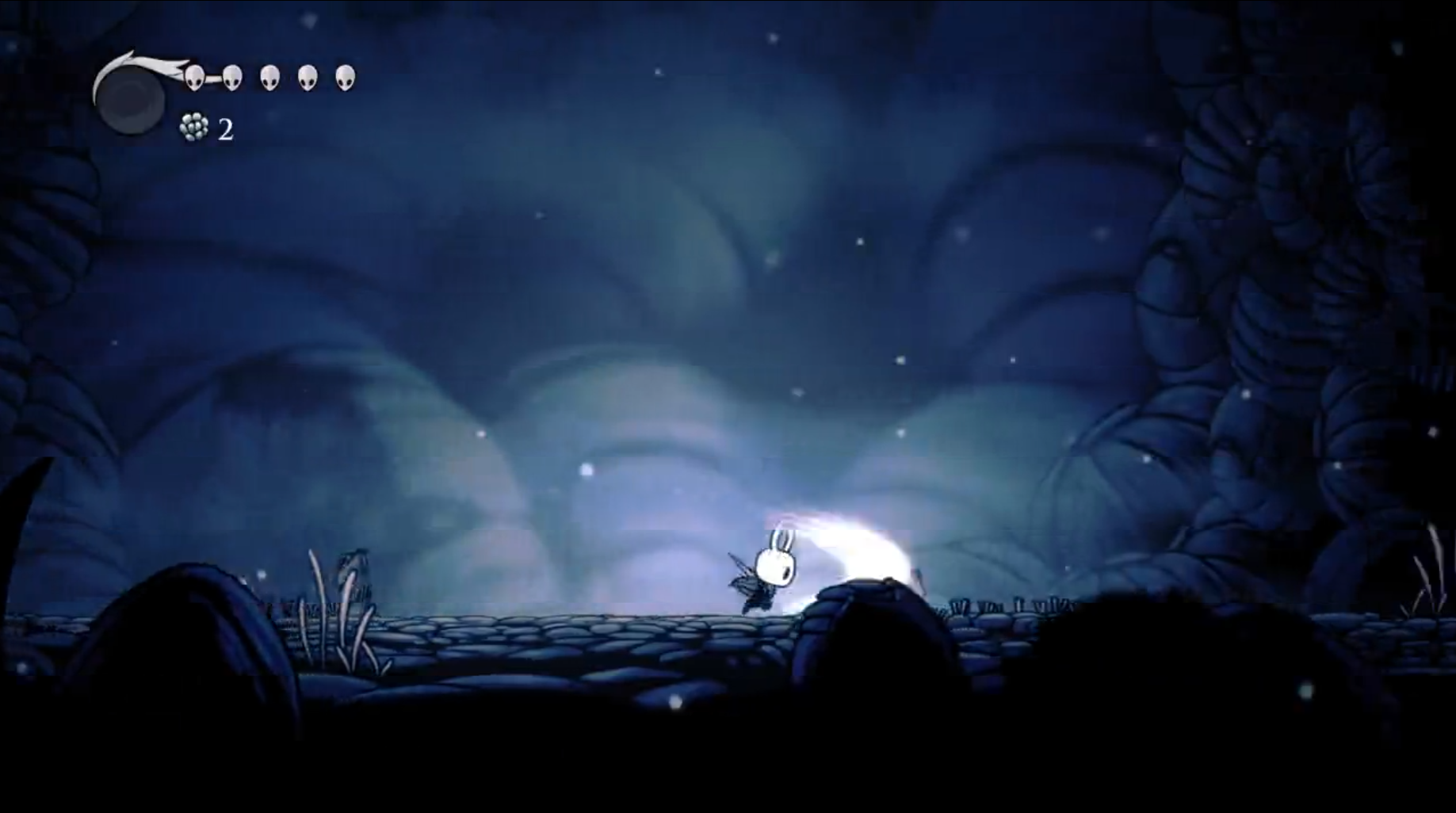}
    \caption{No-guidance condition for visual instruction-following evaluation.}
    \label{fig:d2_no_guidance}
\end{figure}

\begin{figure}[htbp]
    \centering
    \begin{minipage}[t]{0.49\linewidth}
        \vspace{0pt}
        \centering
        \includegraphics[height=0.19\textheight,keepaspectratio]{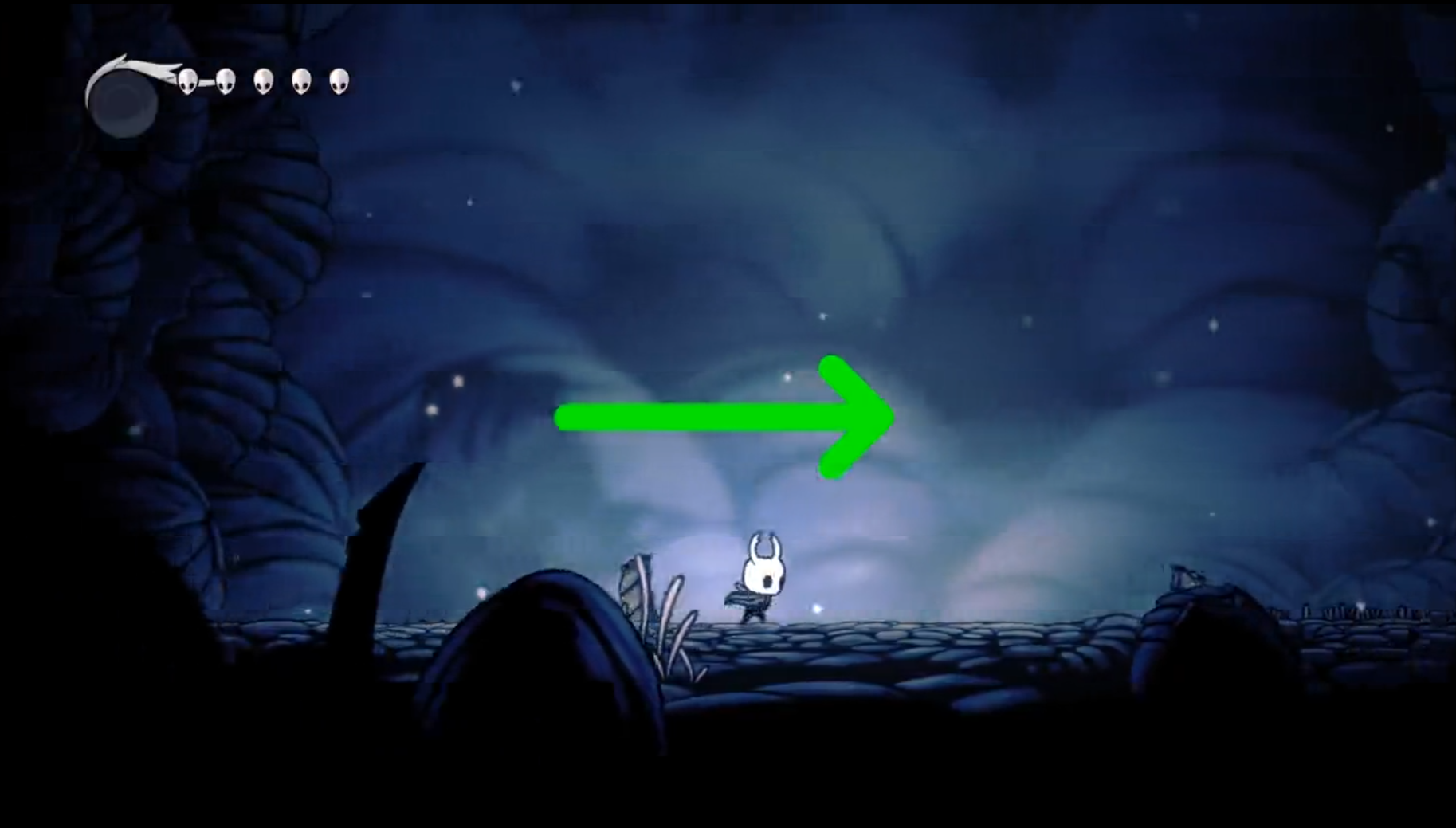}
        \par\smallskip
        \small\textbf{(a)} Graphical guidance
    \end{minipage}
    \hfill
    \begin{minipage}[t]{0.49\linewidth}
        \vspace{0pt}
        \centering
        \includegraphics[height=0.19\textheight,keepaspectratio]{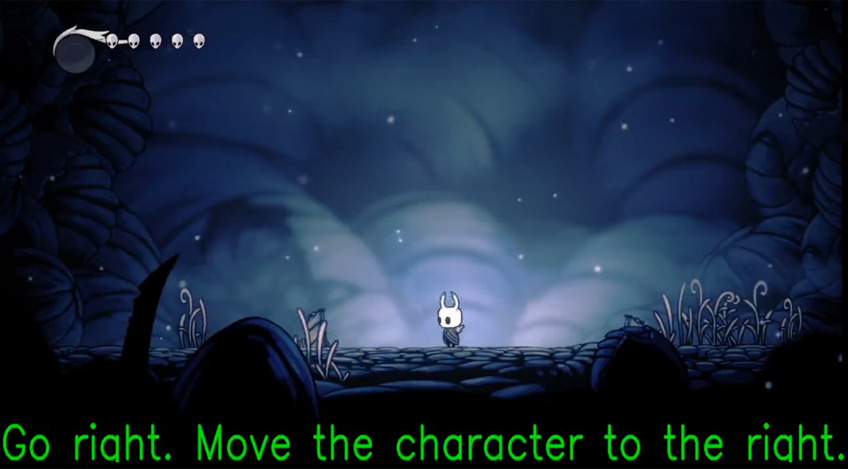}
        \par\smallskip
        \small\textbf{(b)} Text guidance
    \end{minipage}
    \caption{Visual-guidance conditions used in D2: (a) a visible right-arrow overlay and (b) a natural-language instruction embedded in the frame.}
    \label{fig:d2_guidance}
\end{figure}

Tables~\ref{tab:d2_rightward}, \ref{tab:d2_leftward}, and \ref{tab:d2_prob} report the D2 measurements. The arrow prompt does not reliably increase rightward movement. In Hollow Knight, mean rightward movement changes little, from $20.55\%$ to $19.16\%$. In Trackmania, it rises from $1.54\%$ to $23.52\%$. The Hollow Knight effect comes from reduced leftward movement, not added rightward movement. Leftward movement falls from $21.97\%$ to $12.73\%$. The overlay behaves more like a visual perturbation than a directional command. The arrow raises $P(R>L)$ in both games. Even in the best case, Trackmania with the arrow, rightward actions do not dominate in $40\%$ of trials. This measure is not instruction-following accuracy. Text prompts are weaker than graphical arrows in both games.

\begin{table}[htbp]
\centering
\caption{Mean rightward movement per condition ($n=100$ trials).}
\label{tab:d2_rightward}
\vspace{1mm}
\begin{tabular}{lcc}
\toprule
& \multicolumn{2}{c}{Mean Right $\uparrow$} \\
\cmidrule(lr){2-3}
Prompt Type & Hollow Knight & Trackmania \\
\midrule
No Prompt              & 20.55\% & 1.54\%  \\
$\rightarrow$ Arrow    & 19.16\% & 23.52\% \\
``Go Right'' Text      & 13.62\% & 6.88\%  \\
\bottomrule
\end{tabular}
\end{table}

\begin{table}[htbp]
\centering
\caption{Mean leftward movement per condition ($n=100$ trials).}
\label{tab:d2_leftward}
\vspace{1mm}
\begin{tabular}{lcc}
\toprule
& \multicolumn{2}{c}{Mean Left $\downarrow$} \\
\cmidrule(lr){2-3}
Prompt Type & Hollow Knight & Trackmania \\
\midrule
No Prompt              & 21.97\% & 7.18\% \\
$\rightarrow$ Arrow    & 12.73\% & 9.62\% \\
``Go Right'' Text      &  9.35\% & 8.10\% \\
\bottomrule
\end{tabular}
\end{table}

\begin{table}[htbp]
\centering
\caption{Trial-level directional dominance per condition ($n=100$ trials).}
\label{tab:d2_prob}
\vspace{1mm}
\begin{tabular}{lcc}
\toprule
& \multicolumn{2}{c}{$P(R>L)$ $\uparrow$} \\
\cmidrule(lr){2-3}
Prompt Type & Hollow Knight & Trackmania \\
\midrule
No Prompt              & 39\% & 18\% \\
$\rightarrow$ Arrow    & 52\% & 60\% \\
``Go Right'' Text      & 46\% & 34\% \\
\bottomrule
\end{tabular}
\end{table}

The behavior shows partial visual steerability. It does not show full instruction following. The prompt effect is too weak to serve as a command interface. The text prompt is weaker in both games. This fits the model interface: NitroGen maps pixels to actions and does not receive language as a trained command channel. The text conditions also reduce both rightward and leftward movement. The text overlay appears to perturb the scene representation. It does not add clear directional meaning.

Visual prompts are not useless; however, they are unreliable. Prompt response depends on modality and environment. A graphical cue can bias actions in one setting. The same cue does not give robust command following across the tested settings. We describe the failure as weak semantic leverage, not as zero visual influence. Additional D2 results are reported in Appendix~\ref{app:d2}.

\subsection{Spatial Exploration}

\begin{table}[htbp]
\centering
\caption{Spatial entropy and state coverage over fifteen-minute open-space trajectories sampled at 1 FPS. NitroGen visits fewer visual states and has lower entropy than both random and human baselines.}
\label{tab:d3_entropy}
\vspace{1mm}
\begin{tabular}{lcc}
\toprule
Agent & Visited Clusters / 50 & Entropy $H$ \\
\midrule
NitroGen & 18 & \textbf{2.55} \\
Random & 22 & 2.84 \\
Human & 39 & 3.28 \\
\bottomrule
\end{tabular}
\end{table}

\begin{figure}[htbp]
    \centering
    \includegraphics[width=\linewidth]{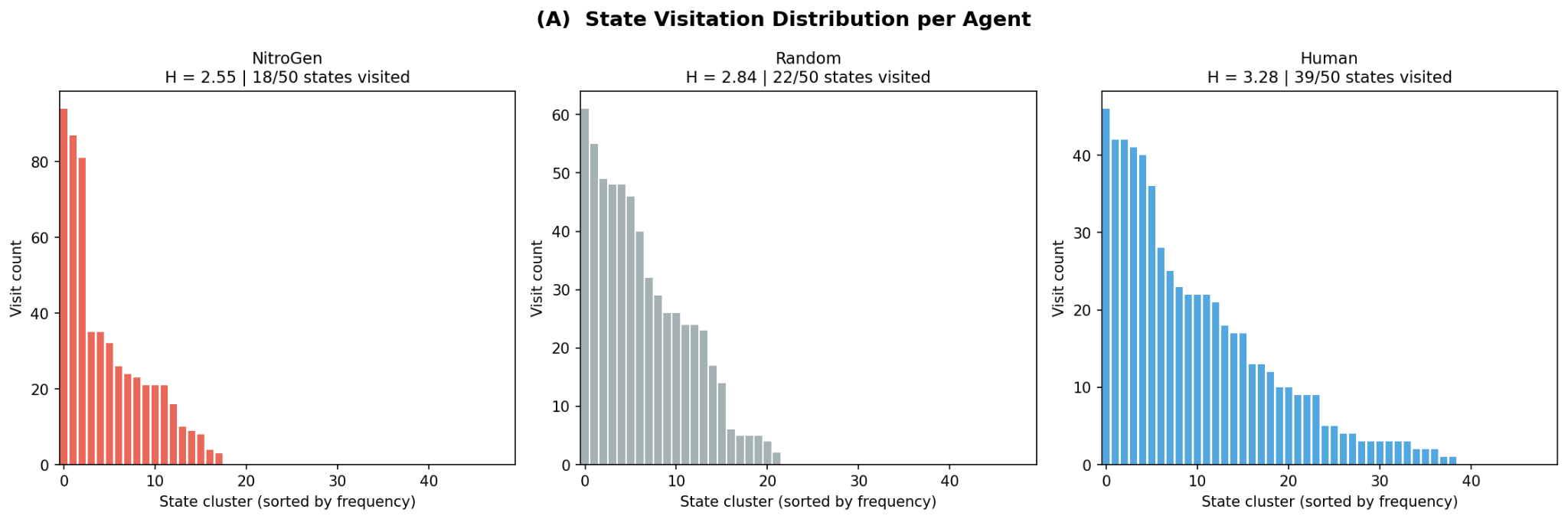}
    \caption{State visitation distribution for fifteen-minute open-space trajectories. NitroGen occupies fewer clusters and concentrates more visits in the most frequent clusters.}
    \label{fig:d3_state_visitation}
\end{figure}

\begin{figure}[htbp]
    \centering
    \includegraphics[width=0.65\linewidth]{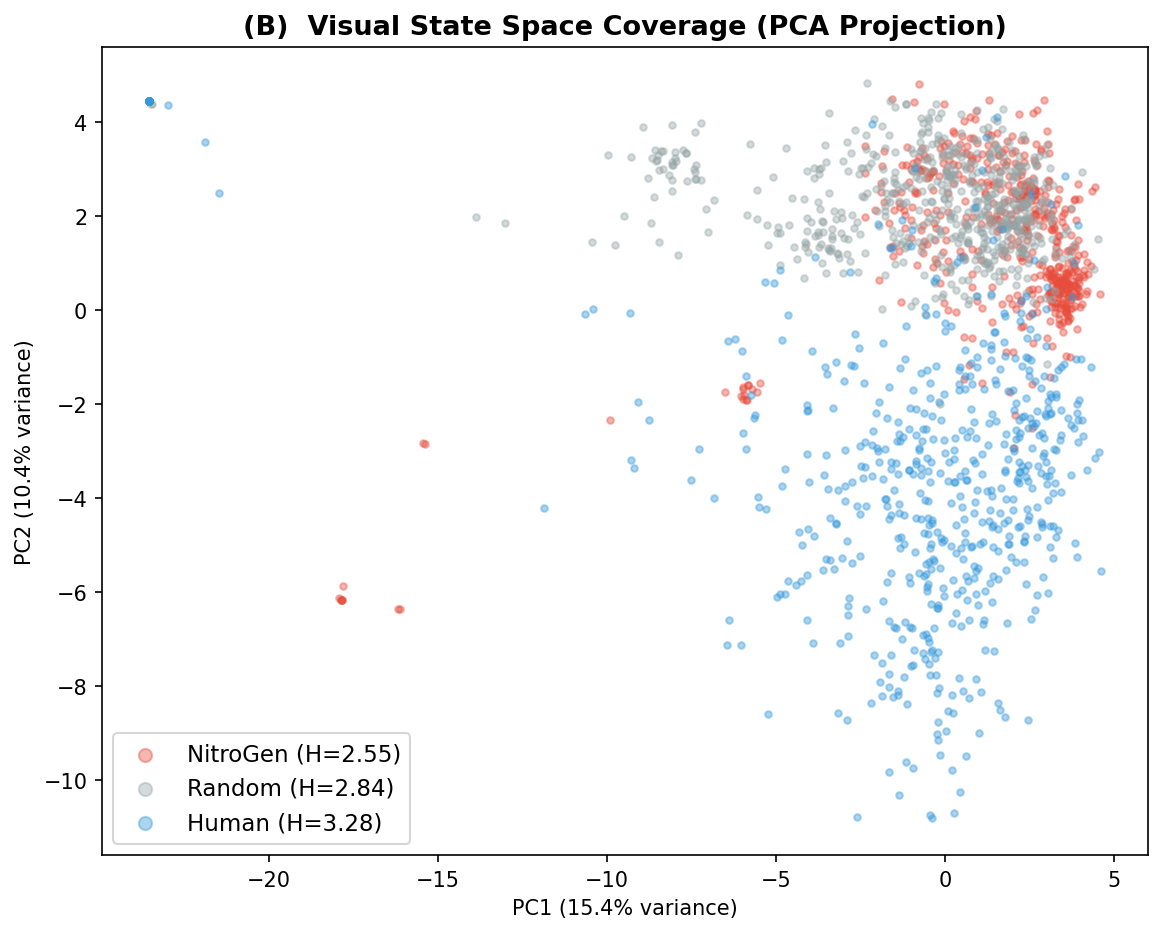}
    \caption{PCA projection of embedded visual states from fifteen-minute open-space trajectories. NitroGen samples a narrower region than the human baseline.}
    \label{fig:d3_pca_projection}
\end{figure}

\begin{figure}[htbp]
    \centering
    \includegraphics[width=0.68\linewidth]{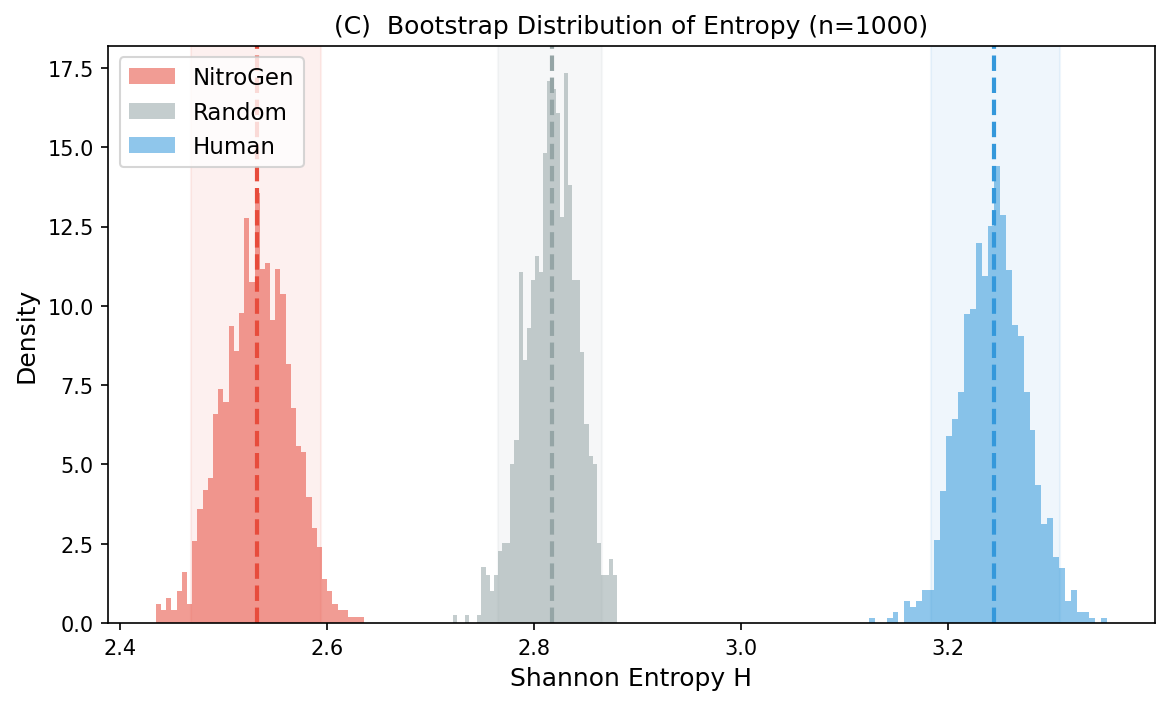}
    \caption{Bootstrap entropy distribution with $n=1000$ resamples from fifteen-minute open-space trajectories. NitroGen remains separated from the random and human baselines.}
    \label{fig:d3_bootstrap_entropy}
\end{figure}

Table~\ref{tab:d3_entropy} reports the entropy and cluster counts for the original D3 recordings. The following figures and effective-state counts also describe these recordings. NitroGen visits 18 of 50 clusters. Random visits 22. Human play visits 39. NitroGen also has the lowest entropy: $H=2.55$, compared with $H=2.84$ for Random and $H=3.28$ for Human. The gap is substantial under the same embedding and clustering pipeline. NitroGen underperforms both Random and Human in measured visual-state coverage. Fig.~\ref{fig:d3_state_visitation} shows the more concentrated NitroGen visitation pattern. Fig.~\ref{fig:d3_pca_projection} shows its narrower region in the visual embedding. Bootstrap resampling with $n=1000$ keeps the entropy distributions visually separated in Fig.~\ref{fig:d3_bootstrap_entropy}. This frame-level bootstrap does not establish the ordering across independent runs. Appendix~\ref{app:d3} reports run-level uncertainty for the additional trajectories. We do not claim exhaustive coverage of all NitroGen environments.

The Random comparison is important. Human play should cover broader and more meaningful states because it reflects purposeful navigation. Random play has no goal. In the full additional dataset, NitroGen visits more clusters than Random but fewer than Human on average. Its mean entropy remains lower than Random's. Coverage and entropy measure different properties, and the coverage ordering changes in the same-resolution check at $K=200$. D1 and D3 are compatible. NitroGen can shift local actions and still remain spatially confined over the full trajectory.

In effective state counts, $K_{\mathrm{eff}}=e^{H}$. NitroGen reaches about $12.8$ effective states. Random reaches about $17.1$. Human play reaches about $26.6$. The human baseline covers roughly twice as many effective visual states as NitroGen.

\section{Discussion: Physical Interpretation of Results}

The three analyses point to the limits of the released controller. NitroGen changes its local action distribution when the game context changes. In D1, it reduces horizontal control and increases jumping after reaching a jump-dependent obstacle. This is positive evidence of local adaptation. It is not evidence of successful long-horizon play because the agent never clears the obstacle.

D2 shows a second boundary. Pixel-level prompts have limited control over the frozen policy. Graphical arrows can shift actions in some settings, but text prompts are weaker. This is expected for a model trained as a pixel-to-action controller rather than as a language-conditioned policy. D3 adds a trajectory-level view. In the full additional dataset, NitroGen visits more visual clusters than Random but fewer than Human on average. Its mean entropy is lower than Random's. The controller can look locally responsive while remaining spatially confined.

\subsection{Relation to Behavior Cloning and Steering}

These findings are consistent with known limits of behavior cloning. A policy can reproduce locally appropriate actions without representing the decision process that generated them. DAgger identifies the distribution-shift problem for cloned policies~\citep{ross2011dagger}. Causal confusion gives a related explanation: an imitation learner may rely on correlations that do not support recovery or goal maintenance~\citep{dehaan2019causal}. Our experiments do not identify the internal cause. They show the external pattern: local adaptation does not reliably produce obstacle completion, prompt controllability, or broad exploration.

This matters for inference-time steering. A high-level module may recognize that the agent is stuck, but the low-level controller still has to execute the correction. Systems such as SayCan separate high-level planning from low-level affordances~\citep{ahn2022saycan}. Language- and task-conditioned policies such as RT-2 and VIMA expose conditioning channels during training~\citep{brohan2023rt2,jiang2023vima}. NitroGen, in the evaluated interface, does not. Text or arrows rendered into the frame are only visual changes. They are not native commands.

\subsection{Scope of the Baselines}

D1 uses each trial as its own paired reference across Early and Late windows. It does not need a random action baseline. Random and Human baselines are used in D3 for visual-state coverage. Human play is a proxy ground truth because NitroGen is trained from human gameplay videos. It is not an exact optimal policy. Random play is not a desirable gameplay strategy either. Its role is diagnostic: if NitroGen covers fewer states than Random under the same embedding and clustering pipeline, then the controller is spatially conservative in the measured trajectories.

\begin{table}[htbp]
\centering
\caption{Scope of conclusions supported by the current evidence.}
\label{tab:scope_conclusions}
\vspace{1mm}
\footnotesize
\begin{tabular}{@{}p{0.42\linewidth}p{0.44\linewidth}@{}}
\toprule
Supported by our experiments & Not established by our experiments \\
\midrule
The tested controller changes its local action distribution with gameplay context but fails to clear the obstacle. & The action shift proves semantic scene understanding or successful planning. \\
Simple visual prompts are not a reliable command channel. & No possible prompt or adapter can steer the model. \\
NitroGen has lower mean entropy than Random in the additional runs. Its entropy ordering relative to Human depends on the cluster count. & State entropy is identical to game progress or task success. \\
Inference-time steering must overcome low-level execution constraints. & High-level reasoning modules are useless in all settings. \\
\bottomrule
\end{tabular}
\end{table}

\subsection{Limitations}

Our study is diagnostic in nature and limited in scale. D1 uses five six-minute Hollow Knight trials from one initial state. The Early and Late windows contain different gameplay contexts. Their difference is not a pure effect of elapsed time. D2 tests one rightward-steering setup. Other prompts may behave differently. Placement, duration, and visual salience may also matter. D3 measures visual-state coverage with a SigLIP--PCA--$k$-means pipeline. It does not measure game progress directly. Entropy values may change with another embedding or clustering method~\citep{guo2021gem}.

The overall evidence is consistent across the three dimensions studied: D1 shows local action adaptation and a task-completion boundary. D2 shows weak prompt controllability. D3 shows restricted visual-state exploration. No single metric in and of itself determines the final conclusion.

\subsection{Broader Impact}

Large-scale online gameplay data may raise copyright and privacy concerns. This diagnostic study does not assess whether the original training data addressed those concerns. The observed control failures also caution against deploying such controllers without safeguards in interactive systems that require high reliability.

\subsection{Future Work}

Future work should separate perception, planning, and control more directly. One option is to keep NitroGen as a low-level motor prior and add an action-space interface for high-level commands. Another option is to train an explicit instruction-conditioned module. Future evaluations should also use controlled pairs of game states, more games, and recovery-oriented scenarios. Exploration analysis should be repeated with other embeddings and clustering schemes to test the robustness of the NitroGen--Random--Human ordering.

\section{Conclusions}

The tested NitroGen controller shows context-sensitive local adaptation. It also shows limited long-horizon execution. Across five Hollow Knight trials, it shifts from horizontal navigation toward more frequent jumping after reaching a jump-dependent obstacle. However, it never clears the obstacle. Visual prompts provide limited steering. In the full additional dataset, NitroGen visits more visual clusters than Random but fewer than Human on average. Its mean entropy is lower than Random's. The Human--NitroGen entropy ordering depends on the cluster count. These results do not establish a fixed ordering of exploration ability. 

The limitations of NitroGen are clear. The frozen controller can select locally relevant action types, but it does not reliably turn those reactions into obstacle completion, external controllability, or broad exploration. While NitroGen remains useful as a reactive visual-motor foundation, future systems might need stronger planning, recovery, and control interfaces. 

\bibliography{main}
\bibliographystyle{tmlr}

\clearpage
\appendix
\renewcommand{\thefigure}{\thesection.\arabic{figure}}
\renewcommand{\thetable}{\thesection.\arabic{table}}
\renewcommand{\theHfigure}{appendix.\thesection.\arabic{figure}}
\renewcommand{\theHtable}{appendix.\thesection.\arabic{table}}
\input{appendices.tex}

\end{document}

%% file: math_commands.tex
\usepackage{amsmath,amsfonts,bm}

\def\eqref#1{equation~\ref{#1}}

\def\1{\bm{1}}

\DeclareMathAlphabet{\mathsfit}{\encodingdefault}{\sfdefault}{m}{sl}
\SetMathAlphabet{\mathsfit}{bold}{\encodingdefault}{\sfdefault}{bx}{n}



%% file: appendices.tex
\section*{APPENDIX}

To test the validity of the main conclusions of our study regarding D1, D2, and D3, additional experiments were conducted. The main findings corroborate the key conclusions reported in the main part of the paper. The details of the extensive experiments we conducted are reported in the following three Appendices.

\section{Additional D1 Experiments}
\label{app:d1}
\setcounter{figure}{0}
\setcounter{table}{0}

\subsection{Setup}

We tested three starting regions in Hollow Knight: Spawn, Corridor, and the jump section. This analysis uses 19 trials with complete local action logs. There are 10 Spawn trials, four Corridor trials, and five jump-section trials. An earlier summary also included 11 externally supplied rows. We exclude those rows here because their full action logs were not available for this analysis.

Each trial contains 21,600 issued control steps. We compare steps 0--3,599 with steps 18,000--21,599. These are the first and last 60 seconds under the nominal 60-Hz control schedule. They are not measurements from an engine-frame counter. Both windows come from the same trial.

We reset the model context before each trial. These trials did not use explicitly paired random seeds. We used a game trainer and recorded routes to reach the starting regions. Position reset was not a full save-state restore. Health, enemies, and prior game history could differ. Exact player coordinates were not matched. This is therefore not a comparison in which only the scene changes while all other state is fixed.

We classify horizontal input as left, right, or no horizontal input. We use the stronger absolute value of normalized left-stick X and net horizontal D-pad input. The stick wins ties. Values below the $\pm0.25$ threshold count as no horizontal input. This does not mean that the character is idle. Jump and attack are the frequencies of the \texttt{SOUTH} and \texttt{WEST} buttons. Movement, jumping, and attacking can occur together.

\subsection{Window Comparison}

\begin{figure}[htbp]
\centering
\includegraphics[width=\linewidth]{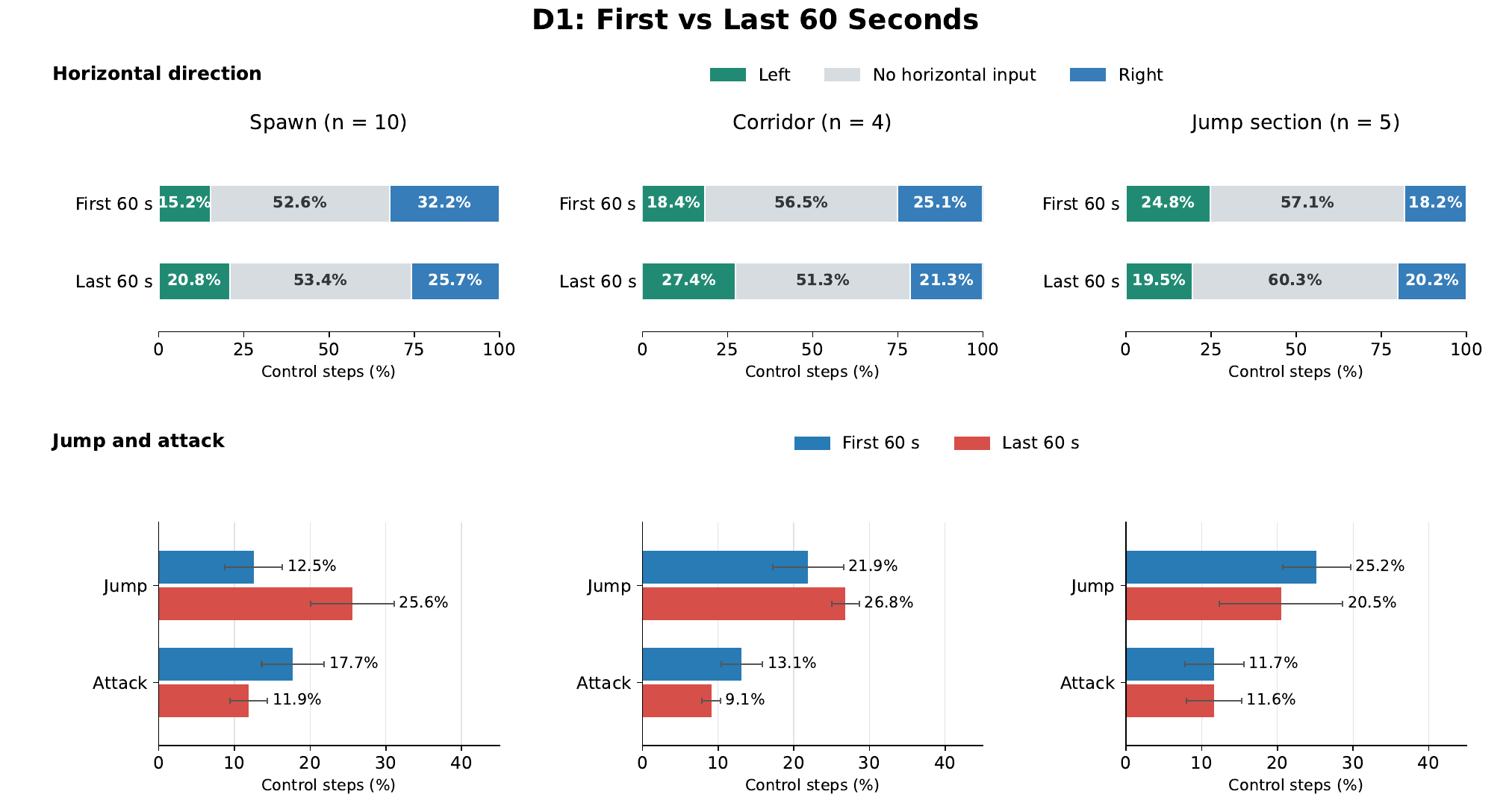}
\caption{D1 actions in the first and last 60 nominal control seconds. Top: mean left, no-horizontal-input, and right proportions. Each trial/window sums to 100\%. Bottom: mean jump and attack frequencies. Whiskers show one sample standard deviation across trials, clipped at 0\% and 100\%. Jump and attack can occur together and during movement. Sample sizes are 10, four, and five for Spawn, Corridor, and the jump section.}
\label{fig:app_d1_actions}
\end{figure}

Figure~\ref{fig:app_d1_actions} shows the action distributions. Table~\ref{tab:app_d1_jump} gives the jump rates. The last-window jump rates from Spawn and Corridor are 25.59\% and 26.82\%. The first-window rate at the jump section is 25.17\%. The similar rates are consistent with a response to the current scene. They do not show that jumping ability improves over time. They also do not rule out every effect of elapsed time or rollout history.

\begin{table}[htbp]
\centering
\caption{Mean jump-button frequency in the 19 locally logged D1 trials. These trials are the same ones used in Figure~\ref{fig:app_d1_actions}.}
\label{tab:app_d1_jump}
\begin{tabular}{lrrr}
\toprule
Starting region & Trials & First 60 s & Last 60 s \\
\midrule
Spawn & 10 & 12.52\% & 25.59\% \\
Corridor & 4 & 21.90\% & 26.82\% \\
Jump section & 5 & 25.17\% & 20.47\% \\
\bottomrule
\end{tabular}

\end{table}

\subsection{A Platform-Jump Action Example}

\begin{figure}[htbp]
\centering
\includegraphics[width=\linewidth]{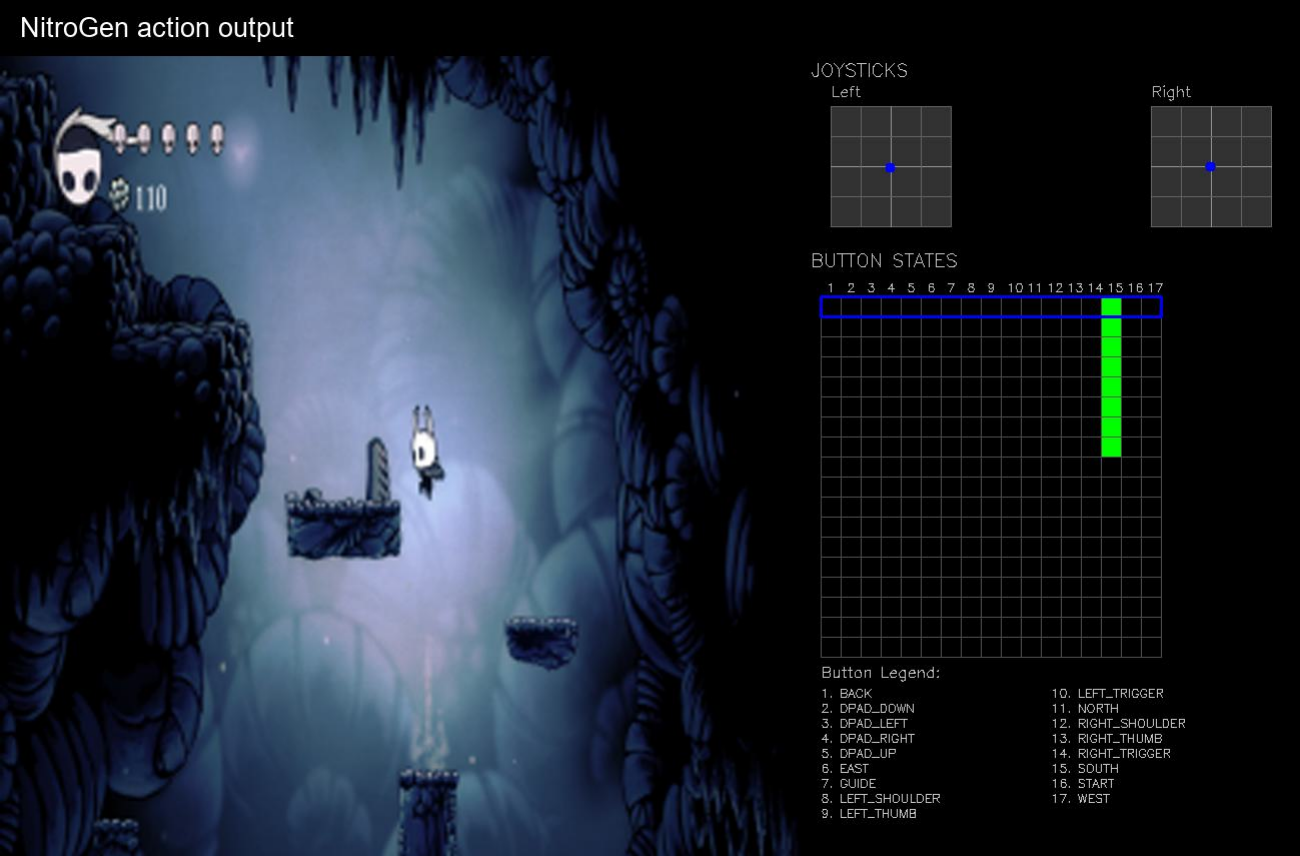}
\caption{Saved input 771 from D1 jump trial 1, followed by its 18 issued commands. This input occurs at 231.3 nominal control seconds. The blue outline marks the first command. Green cells mark active buttons. Column 15 is \texttt{SOUTH}, the jump button. The joystick display shows the first command. Only the 17 logged button channels are shown. The figure uses the runtime visualization layout and is reconstructed from saved data.}
\label{fig:app_d1_output}
\end{figure}

Figure~\ref{fig:app_d1_output} shows a saved input from a jump-section trial and the actions that followed it. The character is above and to the right of the middle floating platform. The first command presses jump but does not issue a substantial horizontal input. Left-stick X is approximately $-0.00275$, well below the movement threshold. Neither horizontal D-pad button is pressed. Horizontal stick values stay within $\pm0.005$ throughout the following 18 commands.

This example shows a lack of horizontal correction at this moment. It does not prove why the model chose these actions. It also does not establish that the whole attempt failed. The right panel shows action outputs, not neural hidden states or attention. We reconstructed it from the saved input and action log. It is not a frame from a debug video.

\clearpage
\section{Additional D2 Experiments}
\label{app:d2}
\setcounter{figure}{0}
\setcounter{table}{0}

\subsection{Directional Cues}

We collected 372 valid Hollow Knight trials. We tested three starting regions and four cues: a right arrow, a left arrow, rightward text, and leftward text. Each region/cue combination contains 31 trials. The left arrow mirrors the right arrow within the same bounding box. The text cues say ``Go right. Move the character to the right.'' and ``Go left. Move the character to the left.'' Green text appears along the bottom of the model input. Each overlay is present in every input throughout the trial.

Each trial contains 600 control steps under a nominal 60-Hz schedule. This corresponds to 10 nominal control seconds. Before each trial, we reset the game and followed a recorded route to the intended starting region. We reset the model context after the route. Each trial uses a distinct recorded model seed. Cue order is randomized within scene/repetition blocks. The routes do not guarantee identical player coordinates or full character states.

Twenty trials used an earlier start-region gate. The other 352 used route-defined starts. We retain both groups in the descriptive summary. One of the 372 valid recordings lacks a completed final reset. A separate check excludes that recording. The direction-reversal intervals still include zero in that check and in the route-only analysis.

\subsection{Movement and Directional Dominance}

We compute left and right action frequencies for each trial. Horizontal direction uses the same combined stick/D-pad rule as Appendix~\ref{app:d1}. We average the trial frequencies within each region/cue combination. We use 10,000 whole-trial bootstrap resamples for their 95\% intervals. We do not treat control steps as independent trials.

We also report $P(R>L)$. This is the fraction of trials with more right than left actions. Ties remain in the denominator but do not count as right-dominant. We use 95\% Wilson intervals for this measure. The definition is the same for left and right cues. It is not instruction-following accuracy.

\begin{figure}[htbp]
\centering
\includegraphics[width=\linewidth]{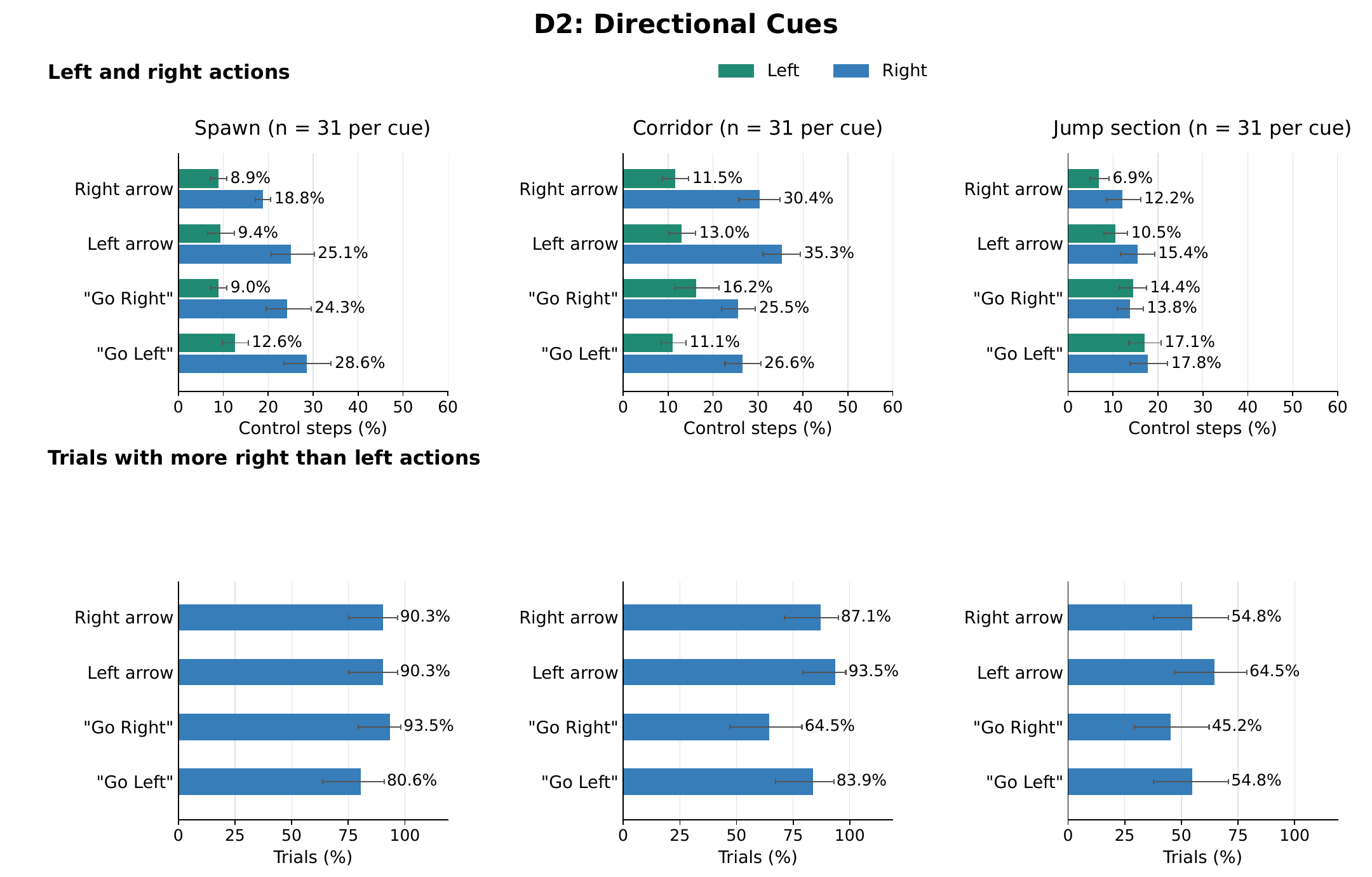}
\caption{D2 directional-cue results. Each region has 31 trials per cue. Top: mean left and right action frequencies with 95\% whole-trial bootstrap intervals. Bottom: the fraction of trials with more right than left actions, with 95\% Wilson intervals. This bottom measure is not cue-following accuracy. Action frequencies are not displacement or task success.}
\label{fig:app_d2_directional}
\end{figure}

Figure~\ref{fig:app_d2_directional} shows both measurements without repeating the movement plot. Left cues did not produce a reliable leftward response. For example, at Spawn, rightward actions average 18.81\% with the right arrow and 25.10\% with the left arrow. Rightward actions also remain more common than leftward actions under both arrows.

We compared the right-minus-left action frequency under right and left cues within blocks. All six region/modality contrast intervals include zero. None passes the Holm-adjusted 0.05 threshold for this six-contrast family. This does not prove that the cues have no effect. It shows that these experiments did not establish reliable direction following.

\subsection{Separate Direction-Free Controls}

We collected 200 complete control trials in a later campaign. Forty trials have no cue. Each of the four pixel-scrambled cues also has 40 trials. Each control condition has 14 Spawn trials, 13 Corridor trials, and 13 jump-section trials. The campaign stopped before the planned 465 trials were complete. We exclude the failed and uncollected trials.

\begin{figure}[htbp]
\centering
\includegraphics[width=\linewidth]{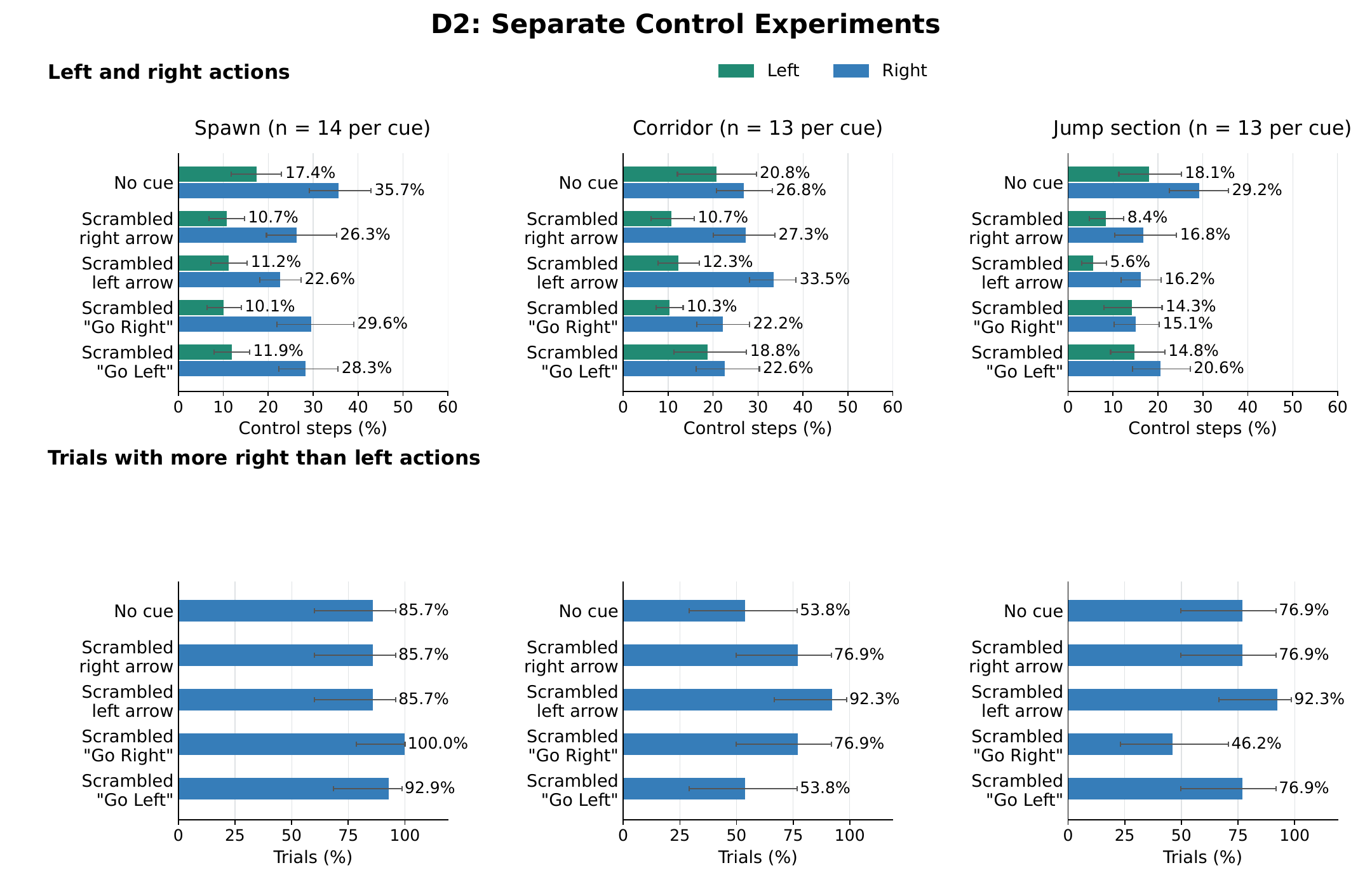}
\caption{Results from the separate 200-trial control campaign. Each condition has 14 Spawn trials, 13 Corridor trials, and 13 jump-section trials. Top: left and right action frequencies with 95\% whole-trial bootstrap intervals. Bottom: rightward dominance with 95\% Wilson intervals. These trials are not contemporaneously paired with Figure~\ref{fig:app_d2_directional}.}
\label{fig:app_d2_controls}
\end{figure}

We later collected 53 more no-cue trials. Together with the original 40, these give 93 no-cue trials, with 31 at each starting region. The original 40 trials are shared between the 200-trial control summary and the 93-trial no-cue summary. They are not counted twice. Figure~\ref{fig:app_d2_controls} reports the original 200-trial control campaign.

The scrambled overlays preserve the source cue's colors, pixel histogram, and bounding box. They do not preserve its spatial layout or spatial frequency. They therefore test a matched set of cue pixels without the original directional arrangement. They do not control every possible visual property.

Each control trial contains 600 control steps and uses a distinct acknowledged model seed. Figure~\ref{fig:app_d2_controls} reports their movement frequencies and directional dominance. We use the same interval methods as for the directional-cue trials. These controls were collected in a separate period. We do not pair them with the earlier 372 trials or combine the campaigns in a causal cue-effect estimate. Comparisons across campaigns are descriptive.

For example, Spawn trials without a cue have mean right and left action frequencies of 35.65\% and 17.36\%. Rightward actions dominate 85.71\% of these trials. A high $P(R>L)$ can therefore occur without a directional instruction.

\clearpage
\section{Additional D3 Experiments}
\label{app:d3}
\setcounter{figure}{0}
\setcounter{table}{0}

\subsection{Trajectories and Human Participants}

The additional dataset contains 13 independent NitroGen trajectories, 14 independent Random trajectories, and four human trajectories. Each human trajectory comes from a different participant. Each trajectory contributes 900 RGB frames sampled at 1 FPS. There are 27,900 frames in total. All frames are processed to $256\times256$ pixels for the visual analysis.

The four participants had experience playing PC games. None had played Hollow Knight before. They were unfamiliar with the tested segment. We recruited them as an intermediate-player baseline, but did not formally test their skill levels. Results may differ for novice or expert players.

All trajectories begin from the common new-game starting point. Before each automated run, we restore the reference save. Each human participant starts a new game before recording. Automated runs use distinct recorded seeds and a 900-second active-time budget. Human recordings use accumulated game-foreground time. These timing records are not a common calibrated engine-frame count.

The pausing regime also differs. Automated runs freeze the game during inference, frame capture, and logging. Human participants play continuously without these pauses. Human recording pauses when the game loses foreground focus. The game may continue during that interval. We therefore treat the human comparison as descriptive.

\subsection{Random Policy and Seeds}

The Random policy selects uniformly from nine action combinations: left, right, left+jump, right+jump, jump, left+attack, right+attack, attack, and no input. Each has probability $1/9$. Each hold duration is independently sampled from Uniform$(0.5,1.0)$ seconds. The policy uses \texttt{numpy.default\_rng(seed)}. Table~\ref{tab:app_d3_seeds} lists the recorded seeds. The model seed controls NitroGen sampling. The policy seed controls Random action and hold-duration sampling.

\begin{table}[htbp]
\centering
\caption{Recorded seeds for the additional automated D3 trajectories. A dash means that no completed NitroGen run is included at that index.}
\label{tab:app_d3_seeds}
\begin{tabular}{rrr}
\toprule
Run & NitroGen seed & Random seed \\
\midrule
1 & 157876371 & 3739848299 \\
2 & 1628221928 & 3082041475 \\
3 & 302451431 & 1420231591 \\
4 & 3014110286 & 3885075862 \\
5 & 2157628747 & 2427701693 \\
6 & 4169213753 & 538775866 \\
7 & 2554724113 & 241840732 \\
8 & 2506167562 & 2120775644 \\
9 & 3047726996 & 3160953415 \\
10 & 2278083733 & 2887840405 \\
11 & 3733536523 & 1643713661 \\
12 & 3072987595 & 3137358625 \\
13 & 2713386701 & 3467964854 \\
14 & --- & 2643545048 \\
\bottomrule
\end{tabular}

\end{table}

\subsection{Shared Visual-State Analysis}

We use the frozen SigLIP visual encoder for all agents. Its frame features have 1,024 dimensions. We fit PCA with 64 dimensions jointly on pooled features from all 31 trajectories. We then fit shared $k$-means clusters with $K=50$, 100, and 200. Each clustering uses seed 42 and 15 initializations. Clustering is not fitted separately for each agent. The PCA projection is the same across the three values of $K$.

We compute three metrics separately for each trajectory: visited-cluster count, Shannon entropy, and the share of frames in its five most visited clusters. Entropy uses natural logarithms. We then average the trajectory metrics within each agent group. A visited cluster is a visual category, not a room or a geographic location.

The native capture resolution was $1920\times1080$ for some automated runs and $1600\times900$ for the others. All human recordings used $1600\times900$. Resizing all inputs to $256\times256$ does not remove every difference in rendering. We therefore report a same-resolution check as well as the full dataset.

\subsection{Results}

\begin{figure}[htbp]
\centering
\includegraphics[width=\linewidth]{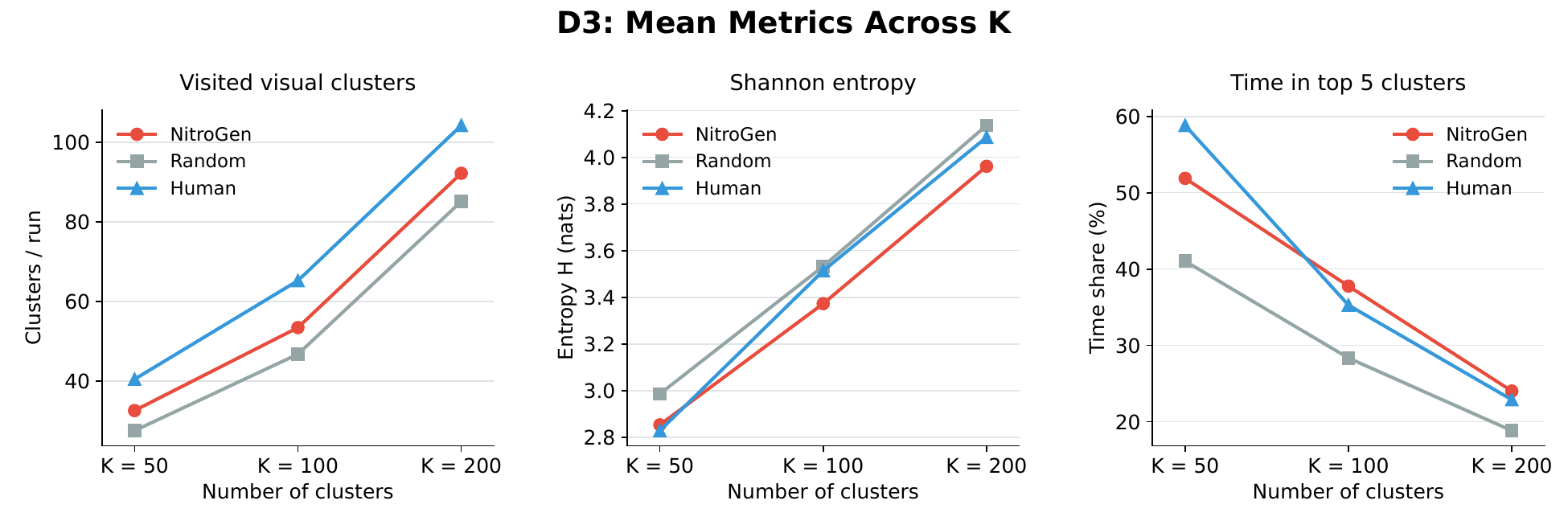}
\caption{Mean visited-cluster count, entropy, and top-five-cluster share in the full D3 dataset. There are 13 NitroGen, 14 Random, and four Human trajectories. Points are means, not confidence intervals. Compare agents within each $K$. Changes across $K$ reflect the visual-state definition, not changes in game progress.}
\label{fig:app_d3_metrics}
\end{figure}

Table~\ref{tab:app_d3_metrics} and Figure~\ref{fig:app_d3_metrics} summarize all 31 trajectories. Human participants have the highest mean visited-cluster count at all three values of $K$. NitroGen visits more clusters on average than Random in this full dataset. NitroGen has lower mean entropy than Random at every $K$. The Human--NitroGen entropy ordering changes with $K$. At $K=50$, the human mean is slightly lower. At $K=100$ and 200, it is higher.

\begin{table}[htbp]
\centering
\caption{Mean D3 metrics across complete trajectories in the full additional dataset. Human values describe the four recorded participants.}
\label{tab:app_d3_metrics}
\begin{tabular}{rlrrrr}
\toprule
$K$ & Agent & Runs & Visited clusters & Entropy $H$ & Top-5 share \\
\midrule
50 & NitroGen & 13 & 32.62 & 2.854 & 51.9\% \\
50 & Random & 14 & 27.57 & 2.987 & 41.0\% \\
50 & Human & 4 & 40.50 & 2.829 & 58.8\% \\
\midrule
100 & NitroGen & 13 & 53.46 & 3.373 & 37.8\% \\
100 & Random & 14 & 46.79 & 3.533 & 28.3\% \\
100 & Human & 4 & 65.25 & 3.514 & 35.3\% \\
\midrule
200 & NitroGen & 13 & 92.23 & 3.962 & 24.0\% \\
200 & Random & 14 & 85.14 & 4.137 & 18.8\% \\
200 & Human & 4 & 104.25 & 4.086 & 22.9\% \\
\bottomrule
\end{tabular}

\end{table}

\begin{figure}[htbp]
\centering
\includegraphics[width=\linewidth]{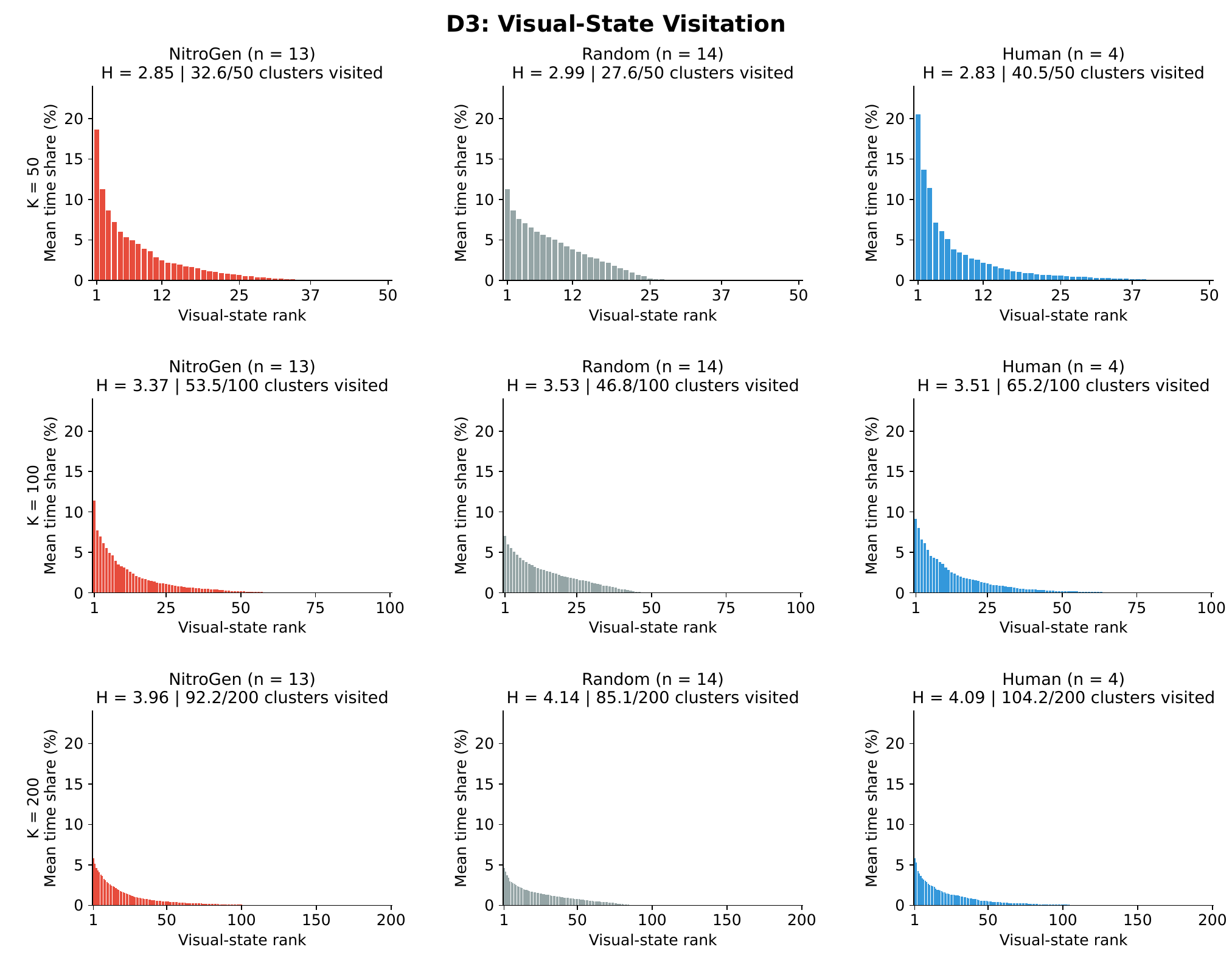}
\caption{D3 visual-state visitation at $K=50$, 100, and 200. Each trajectory is normalized and sorted before averaging within an agent group. All panels use the same vertical scale. Titles give mean per-trajectory entropy and visited-cluster count. State rank does not align cluster identities across agents. Higher $K$ divides the same frames into finer categories; it is not a new gameplay experiment.}
\label{fig:app_d3_visitation}
\end{figure}

Figure~\ref{fig:app_d3_visitation} shows how visits are distributed. We normalize and sort each trajectory's cluster counts before averaging. The horizontal axis is a frequency rank. It is not a shared cluster identity across agents. The plotted curve's entropy is not the mean of the trajectory entropies.

\subsection{Run-Level Uncertainty and Limits}

\begin{figure}[htbp]
\centering
\includegraphics[width=\linewidth]{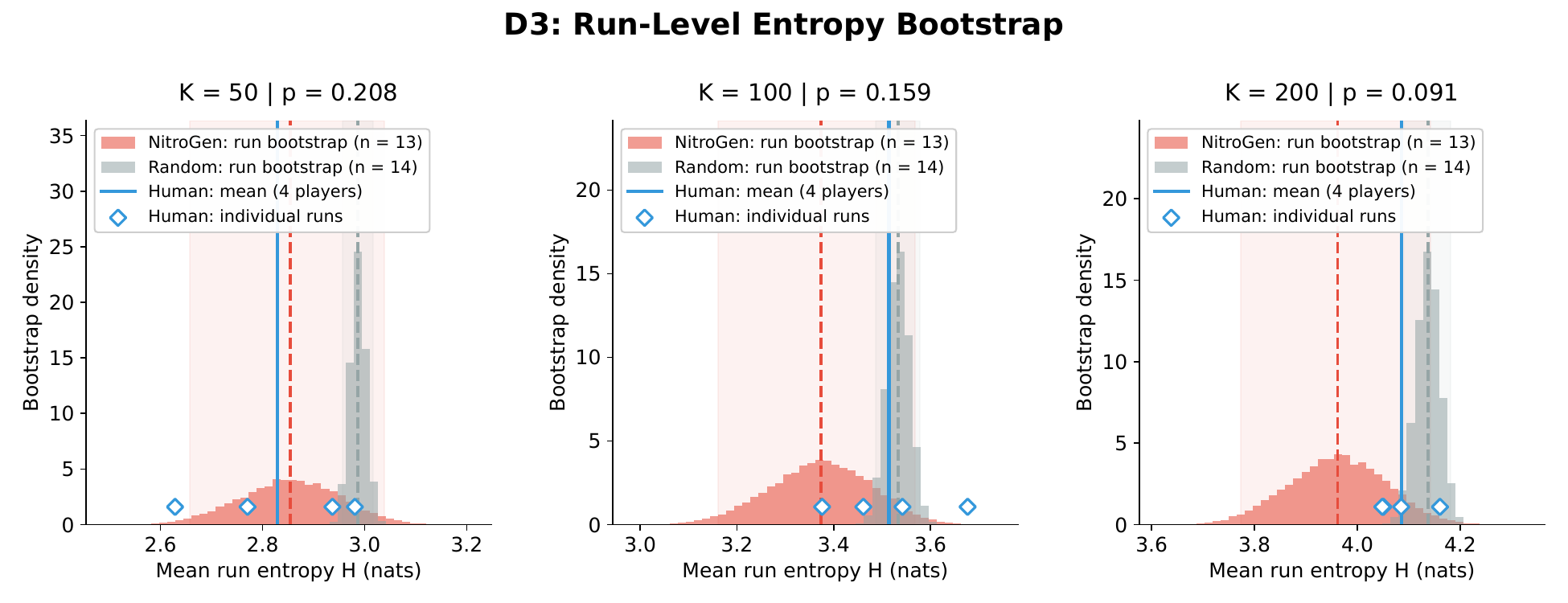}
\caption{Mean-entropy bootstrap from 20,000 resamples of complete automated trajectories. Dashed lines mark automated-group means. Shaded bands mark their conditional 95\% bootstrap intervals. Blue diamonds show the four observed human entropies; the solid blue line marks their mean. Each panel uses its own fixed shared clusters. Reported $p$-values are unadjusted two-sided run-level permutation results.}
\label{fig:app_d3_bootstrap}
\end{figure}

For NitroGen and Random, we bootstrap complete trajectories 20,000 times. Each sampled trajectory keeps all its frames. Figure~\ref{fig:app_d3_bootstrap} shows the bootstrap distribution of mean entropy. These intervals use fixed PCA and cluster assignments. They measure run variation under that visual-state definition. They do not include all uncertainty from fitting the clusters. We show individual human values and their mean, not a population-level human interval.

The full-dataset two-sided run-level permutation $p$-values for NitroGen--Random entropy are 0.208, 0.159, and 0.091 at $K=50$, 100, and 200. None is below 0.05. These are raw, unadjusted $p$-values. The comparisons are exploratory. The three values of $K$ reuse the same trajectories and are not independent replications.

The $1600\times900$ check contains five NitroGen, seven Random, and four Human trajectories. NitroGen's mean visited-cluster counts are 29.40, 47.20, and 80.60. Random's are 27.71, 46.14, and 84.29. Human's are 40.50, 65.25, and 104.25. Thus the NitroGen--Random coverage ordering changes at $K=200$ in this check. NitroGen still has lower mean entropy than Random at all three values of $K$. Its mean entropies are 2.62, 3.13, and 3.73, compared with Random's 2.97, 3.50, and 4.10.

The new results partly support the original D3 observation of concentrated NitroGen visitation. They do not support a fixed ordering of exploration ability across all three agents. Visited-cluster count and entropy measure different properties. A trajectory can briefly visit many clusters but spend most of its time in a few. Human recordings also contain maps, dialogue, and other interface states. These can increase visual coverage without indicating new geographic areas. Different timing and pausing procedures further limit the comparison. More matched runs and direct measures of game progress are needed for broader claims.